\documentclass[runningheads,a4paper]{llncs}
\usepackage[top=2.5cm, bottom=2.5cm, left=2.5cm, right=2.5cm]{geometry}
\usepackage{amsmath}
\usepackage{amssymb}
\usepackage{graphicx}
\usepackage{booktabs}
\usepackage{listings}
\usepackage{xcolor}
\usepackage{hyperref}
\usepackage{caption}
\usepackage{subcaption}

\usepackage{url}
\usepackage{color}
\usepackage{cite}
\usepackage{epsfig}

\usepackage[nomargin,inline,marginclue,draft]{fixme}
\usepackage{cleveref}
\fxsetup{status=draft}
\fxsetup{theme=color, mode=multiuser} \FXRegisterAuthor{me}{ame}{\color{red}Me}
\newcommand{\orcid}[1]{\href{https://orcid.org/#1}{\includegraphics[scale=0.02]{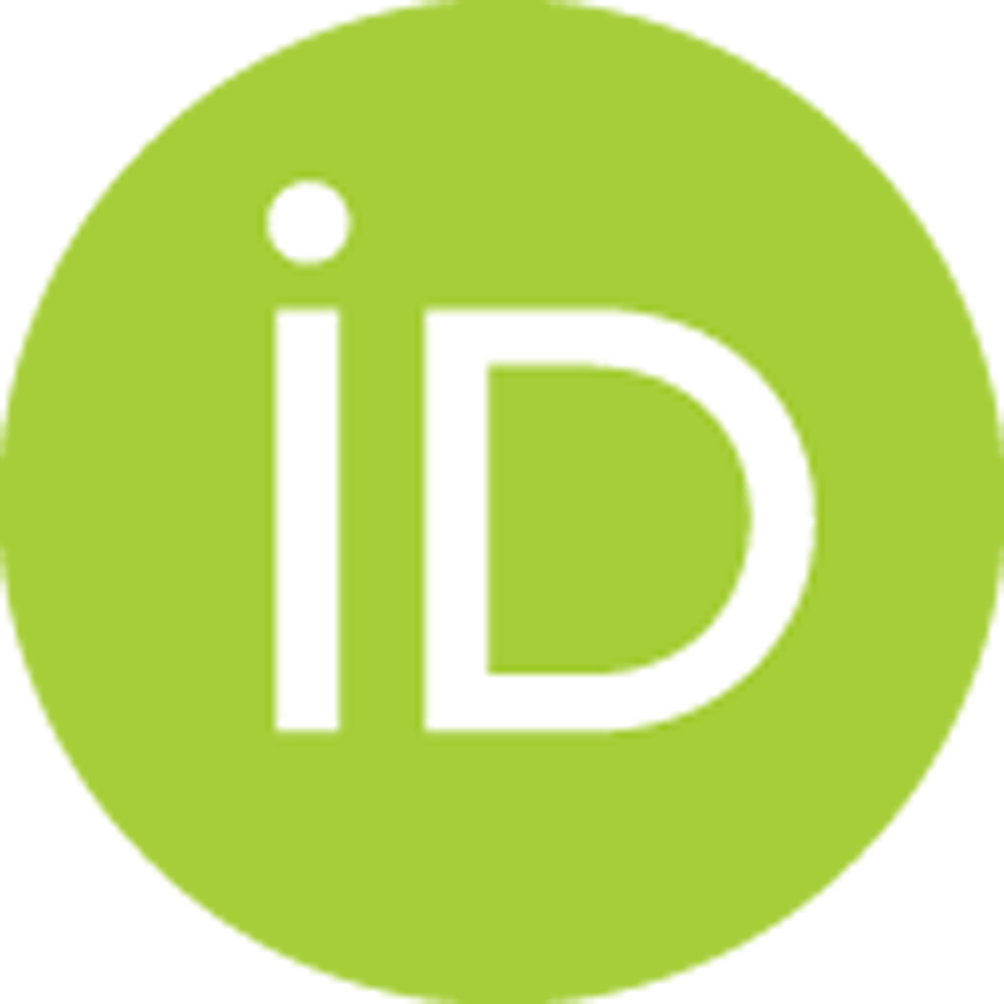}}} 

\hypersetup{
    colorlinks=true
}
\title{The ASNR-MICCAI Brain Tumor Segmentation (BraTS) Challenge 2023: Intracranial Meningioma}

\titlerunning{BraTS 2023 Meningioma Challenge}

\author{
Dominic LaBella\inst{1,\dag,\ddag,\S,\P,*,\orcid{0000-0003-1713-9538}}
\and Maruf Adewole\inst{2,\dag}
\and Michelle Alonso-Basanta\inst{3,\dag,\ddag,\S}
\and Talissa Altes\inst{4,\dag}
\and Syed Muhammad Anwar\inst{5,6,\dag}
\and Ujjwal Baid\inst{7,8,9,\dag,\ddag,\S}
\and Timothy Bergquist\inst{10,\dag}
\and Radhika Bhalerao\inst{11,\dag}
\and Sully Chen\inst{12,\dag}
\and Verena Chung\inst{10,\dag}
\and Gian-Marco Conte\inst{13,\dag}
\and Farouk Dako\inst{14,\dag}
\and James Eddy\inst{10,\dag}
\and Ivan Ezhov\inst{15,16,\dag}
\and Devon Godfrey\inst{1,\dag}
\and Fathi Hilal\inst{4,\ddag}
\and Ariana Familiar\inst{17,\dag}
\and Keyvan Farahani\inst{18,\dag}
\and Juan Eugenio Iglesias\inst{19,\dag}
\and Zhifan Jiang\inst{5,6,\dag}
\and Elaine Johanson\inst{20,\dag}
\and Anahita Fathi Kazerooni\inst{7,17,\dag}
\and Collin Kent\inst{1,\dag}
\and John Kirkpatrick\inst{1,\ddag}
\and Florian Kofler\inst{15,16,21,22,\dag}
\and Koen Van Leemput\inst{23,\dag}
\and Hongwei Bran Li\inst{19,24,25,\dag}
\and Xinyang Liu\inst{5,6,\dag}
\and Aria Mahtabfar\inst{26,\dag, \ddag,\S}
\and Shan McBurney-Lin\inst{11,\ddag}
\and Ryan McLean\inst{27,\dag,\ddag,\S}
\and Zeke Meier\inst{28,\dag}
\and Ahmed W Moawad\inst{29,\dag}
\and John Mongan\inst{11,\ddag}
\and Pierre Nedelec\inst{11,\ddag,\S}
\and Maxence Pajot\inst{11,\ddag,\S}
\and Marie Piraud\inst{21,\dag}
\and Arif Rashid\inst{3,\ddag}
\and Zachary Reitman\inst{1,\dag,\ddag}
\and Russell Takeshi Shinohara\inst{7,30,\dag}
\and Yury Velichko\inst{31,\dag}
\and Chunhao Wang\inst{1,\dag,\ddag}
\and Pranav Warman\inst{12,\dag}
\and Walter Wiggins\inst{32,\dag,\ddag}
\and Mariam Aboian\inst{27,\dag,\ddag,\S,\P}
\and Jake Albrecht\inst{10,\dag,\P}
\and Udunna Anazodo\inst{33,\dag,\P}
\and Spyridon Bakas\inst{7,8,9,\dag,\ddag,\S,\P}
\and Adam Flanders\inst{26,\dag,\P}
\and Anastasia Janas\inst{27,\dag,\S,\P}
\and Goldey Khanna\inst{26,\dag,\ddag,\S,\P}
\and Marius George Linguraru\inst{5,6,\dag,\P}
\and Bjoern Menze\inst{24,25,\dag,\P}
\and Ayman Nada\inst{4,\dag,\ddag,\S,\P}
\and Andreas M Rauschecker\inst{11,\dag,\ddag,\S,\P}
\and Jeff Rudie\inst{11,34,\dag,\S,\P}
\and Nourel Hoda Tahon\inst{4,\dag,\ddag,\S,\P}
\and Javier Villanueva-Meyer\inst{11,\dag,\ddag,\S,\P}
\and Benedikt Wiestler\inst{24,\dag,\P}
\and Evan Calabrese\inst{32,\dag,\ddag,\S,\P,**}
}

\authorrunning{LaBella, et al.}

\institute{\scriptsize{
Department of Radiation Oncology, Duke University Medical Center, Durham, NC, USA
\and Medical Artificial Intelligence (MAI) Lab, Crestview Radiology, Lagos, Nigeria
\and Department of Radiation Oncology, Perelman School of Medicine, University of Pennsylvania, Philadelphia, PA, USA
\and Missouri University, Columbia, MO, USA 
\and Children's National Hospital, Washington DC, USA
\and George Washington University, Washington DC, USA
\and Center for AI and Data Science for Integrated Diagnostics (AI2D) and Center for Biomedical Image Computing and Analytics (CBICA), University of Pennsylvania, Philadelphia, PA, USA
\and Department of Radiology, Perelman School of Medicine, University of Pennsylvania, Philadelphia, PA, USA
\and Department of Pathology and Laboratory Medicine, Perelman School of Medicine, University of Pennsylvania, Philadelphia, PA, USA
\and Sage Bionetworks, USA
\and University of California San Francisco, CA, USA
\and Duke University Medical Center, School of Medicine, USA
\and Mayo Clinic, MN, USA
\and Center for Global Health, Perelman School of Medicine, University of Pennsylvania, Philadelphia, Pennsylvania, USA
\and Department of Informatics, Technical University Munich, Germany
\and TranslaTUM - Central Institute for Translational Cancer Research, Technical University of Munich, Germany
\and Children’s Hospital of Philadelphia, University of Pennsylvania, Philadelphia, PA, USA
\and Cancer Imaging Program, National Cancer Institute, National Institutes of Health, Bethesda, MD 20814, USA
\and Athinoula A Martinos Center for Biomedical Imaging, Massachusetts General Hospital, Boston, MA, USA
\and PrecisionFDA, U.S. Food and Drug Administration, Silver Spring, MD, USA
\and Helmholtz AI, Helmholtz Munich, Germany
\and Department of Diagnostic and Interventional Neuroradiology, School of Medicine, Klinikum rechts der Isar, Technical University of Munich, Germany
\and Department of Applied Mathematics and Computer Science, Technical University of Denmark, Denmark
\and Department of Neuroradiology, Technical University of Munich, Munich, Germany
\and University of Zurich, Switzerland
\and Department of Neurosurgery, Thomas Jefferson University, Philadelphia, PA, USA 
\and Yale University, New Haven, CT, USA
\and Booz Allen Hamilton, McLean, VA, USA
\and Mercy Catholic Medical Center, Darby, PA, USA
\and Center for Clinical Epidemiology and Biostatistics, University of Pennsylvania, Philadelphia, USA
\and Department of Radiology, Northwestern University, Evanston, IL, USA
\and Department of Radiology, Duke University Medical Center, Durham, NC, USA 
\and Montreal Neurological Institute (MNI), McGill University, Montreal, QC, Canada
\and University of San Diego, CA, USA
}
\linebreak
\\
\textsuperscript{\dag} People involved in the organization of the challenge.\\
\textsuperscript{\ddag} People contributing data from their institutions.\\
\textsuperscript{\S} People involved in annotation process.\\ 
\textsuperscript{*} First author.\\
\textsuperscript{\P} Senior authors.\\ 
\textsuperscript{**} Corresponding author: \email{\{evan.calabrese@duke.edu\}}}

\begin{document}
    \mainmatter
    \maketitle
    \setcounter{footnote}{0}
    \begin{abstract}
    Meningiomas are the most common primary intracranial tumor in adults and can be associated with significant morbidity and mortality. Radiologists, neurosurgeons, neuro-oncologists, and radiation oncologists rely on multiparametric MRI (mpMRI) for diagnosis, treatment planning, and longitudinal treatment monitoring; yet automated, objective, and quantitative tools for non-invasive assessment of meningiomas on mpMRI are lacking. The BraTS meningioma 2023 challenge will provide a community standard and benchmark for state-of-the-art automated intracranial meningioma segmentation models based on the largest expert annotated multilabel meningioma mpMRI dataset to date. Challenge competitors will develop automated segmentation models to predict three distinct meningioma sub-regions on MRI including enhancing tumor, non-enhancing tumor core, and surrounding nonenhancing T2/FLAIR hyperintensity. Models will be evaluated on separate validation and held-out test datasets using standardized metrics utilized across the BraTS 2023 series of challenges including the Dice similarity coefficient and Hausdorff distance. The models developed during the course of this challenge will aid in incorporation of automated meningioma MRI segmentation into clinical practice, which will ultimately improve care of patients with meningioma.
    \end{abstract}
    
    \keywords{BraTS, challenge, brain, tumor, segmentation, machine learning, deep learning, artificial intelligence, AI, meningioma}
    
    \section{Introduction}

Meningiomas are the most common primary intracranial tumor in adults and can result in significant morbidity and mortality for affected patients \cite{ogasawara2021meningioma,huntoon2020meningioma}. Most meningiomas ($\sim$80\%) are World Health Organization (WHO) grade 1 benign tumors and are typically well controlled with observation, surgical resection, and/or radiation therapy. However, higher grade meningiomas (WHO grades 2 and 3) are associated with significantly higher morbidity and mortality rates and often recur despite optimal management. Currently there is no reliable noninvasive method for identifying meningioma grade, assessing aggressiveness, or predicting recurrence and survival. Traditional MRI features used by clinicians to guide treatment strategy, such as meningioma volume, or degree of peritumoral edema, may not represent tumor WHO grade, or expected clinical course. As such, there is a need for improved radiographic assessment of meningiomas, such that it can guide patient-specific treatment strategy.

Automated segmentation on brain magnetic resonance imaging (MRI) has matured into a clinically viable tool that can provide objective assessments of tumor volume and can assist in surgical planning, radiotherapy planning, and treatment response assessment. However, to date most tumor segmentation studies have focused on gliomas. Meningiomas, while typically more circumscribed than gliomas, provide additional technical challenges for segmentation given their extra-axial location, multiplicity, and propensity for skull-base involvement. In addition, unlike other intracranial tumors, meningiomas are commonly diagnosed by imaging alone, which increases the importance of MRI for treatment planning. The ability to predict meningioma aggressiveness pre-operatively would confer the benefit of guiding surgical strategy, as maximal resection remains the mainstay of effective treatment to mitigate the risk of recurrence\cite{khanna2021machine}.

The purpose of the Brain Tumor Segmentation (BraTS) 2023 meningioma challenge is to develop an automated multi-compartment brain MRI segmentation algorithm for intracranial meningiomas. This algorithm, if successful, will provide an important tool for objective assessment of tumor volume for surgical and radiotherapy planning. In addition, this algorithm will provide a starting point for future studies focused on identifying meningioma grade, assessing aggressiveness, and predicting risk of recurrence based on MRI findings alone. This manuscript describes the annotation protocol used to prepare the data used in the BraTS 2023 meningioma challenge, and it outlines the challenge tasks and the evaluation metrics used. Additionally, the manuscript explores the advantages, objectives, and constraints of the challenge, and outlines future directions currently under consideration.

    \section{Materials \& Methods}
        \subsection{Defining intracranial meningioma}

For the purposes of this challenge, intracranial meningioma was defined as any grade or subtype of meningioma occurring within the cranial vault. Meningiomas arise from the arachnoid layer of the meninges between the dura mater and pia mater. Intracranial meningiomas commonly present at supratentorial sites of dural reflection, along the sphenoid sinus, and the skull base. Less commonly, meningiomas occur in intraventricular and suprasellar regions, the olfactory groove, and in the posterior fossa along the petrous bone. Approximately 80\% of meningioma are classified as WHO grade 1 and include meningothelial, fibroblastic, transitional, psammomatous, angiomatous, microcystic, secretory, metaplastic, and lymphoplasmacyte rich subtypes. Histologically, WHO grade 1 meningiomas typically have psammoma bodies, cellular whirls and calcifications. Approximately 18\% of meningiomas are classified as WHO grade 2 and include atypical, chordoid, clear cell, and subtypes. Histologically, WHO grade 2 meningiomas have $\geq$ 4 mitoses per high-power field (HPF), brain invasion, or $\geq$ 3 of the following: hypercellularity, small cells, prominent nucleoli, patternless or sheet like growth, foci of spontaneous necrosis on hematoxylin and eosin staining. Only about 2\% of meningiomas are classified as WHO grade 3 and include anaplastic, papillary, and rhabdoid subtypes. Histologically, WHO grade 3 meningiomas have $\geq$ 10 mitoses/HPF and/or carcinomatous features, sarcomatous features, melanomatous features, loss of usual growth pattern, or multifocal necrotic foci. Radiographic findings vary greatly amongst meningioma subtypes, and to date, there are no reliable imaging features that can accurately distinguish meningioma grade or aggressiveness.

\subsection{Defining Meningioma MRI Sub-Components}     

Similar to prior BraTS glioma challenges \cite{bakas2018identifying, menze2014multimodal, bakas2017advancing, bakas2017segmentation_1, bakas2017segmentation_2}, a key aspect of the challenge is the subdivision of the different tumor compartments that are apparent on MRI \cite{bakas2022university, baid2021rsna,bakas2018identifying,baid2020novel}. To date, there are no published guidelines about the specific target volume delineation for intracranial meningioma based on prospective studies \cite{martz2022anocef}. However, many different meningioma MRI appearances and sub-regions have been previously described.  In 2022, Association des Neuro-oncologues d'Expression Française (ANOCEF) outlined consensus guidelines for meningioma gross tumor volume after 20 experts from 17 radiotherapy centers participated in a three round modified Delphi consensus \cite{martz2022anocef,nasa2021delphi}. The ANOCEF committee defined the enhancing gross tumor to include MRI T1 contrast-enhancing lesions, thickened meninges, and directly invaded bone.

Watt et al. described that meningioma non-enhancing calcification is consistent with low signal intensity on T2 imaging \cite{watts2014magnetic}. The microcystic meningioma subtype comprises 1.6\% of meningiomas and demonstrates a uniform low-signal intensity on T1 and high signal intensity of T2. Bitzer et al found that peritumoral edema is found in 60\% of meningioma cases and is demonstrated by a hyperintense FLAIR signal \cite{bitzer1998angiogenesis}. En-plaque meningioma is described as asymmetric thickened sheets of enhancing dura (Fig. 1) \cite{watts2014magnetic}. Dural tail involvement is defined as thickening and enhancement of the dura infiltrating away from the lesion (Fig. 2) and is used to help radiographically distinguish meningioma from other enhancing CNS lesions owing to its high prevalence in meningioma (\(\sim \)72\%) and absence in meningioma mimics like intracranial Schwannomas \cite{watts2014magnetic, o2007atypical,dorent2023crossmoda}.

    \begin{figure}[h]
          \centering
          \includegraphics[width=0.4\columnwidth]{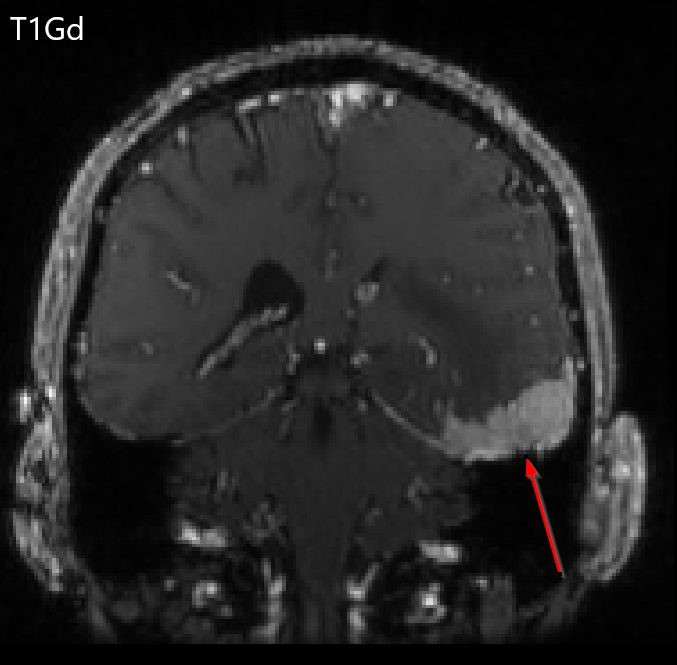}
          \caption{Example of an en plaque meningioma on a T1Gd coronal image.}
        \label{annotations}
    \end{figure}

\begin{figure}[h]
          \centering
          \includegraphics[width=0.4\columnwidth]{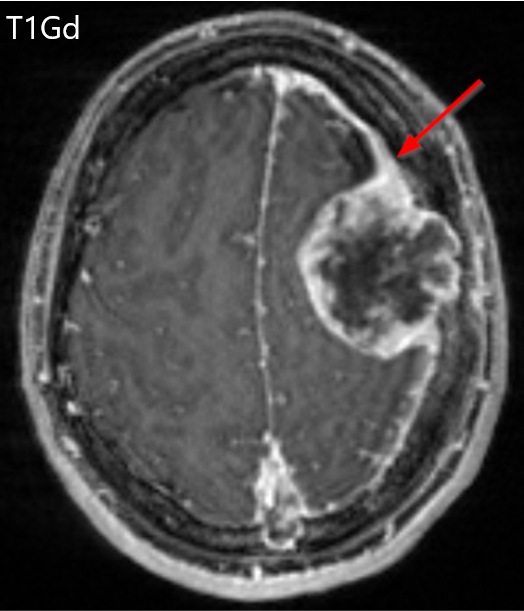}
          \caption{Example of a meningioma dural tail on a T1Gd axial image.}
        \label{annotations}
    \end{figure}

Based on this prior work and others, the BraTS 2023 meningioma challenge defines 3 distinct and non-overlapping segmentation labels (Fig. 3). These include “enhancing tumor”, “nonenhancing tumor core”, and surrounding non-enhancing T2/FLAIR hyperintensity (SNFH). The enhancing tumor label includes all contrast enhancing meningioma, focally thickened meninges (including dural tail), as well as en-plaque meningiomas. This label approximates the region of active, viable tumor. The non-enhancing tumor core label includes all calcification, hyperostosis, necrosis, degeneration, and any other atypical non-enhancing tumor radiographic findings. This label along with the enhancing tumor label (together comprising the “tumor core”) approximately corresponds to the portion of tumor related imaging abnormality that would typically be removed in a gross total resection. The SNFH label includes the entire extent of tumor related T2/FLAIR hyperintensity surrounding the tumor core. This label is distinct from the other labels in that it is composed entirely of brain parenchyma and is not expected to contain any tumor cells, but rather represents irritated, inflamed, and/or edematous brain tissue resulting from adjacent tumor. Importantly, non-tumor related T2/FLAIR signal abnormality, commonly related to chronic microvascular ischemic white matter changes (e.g. leukoaraiosis) or other vascular pathology, was not included in the SNFH label.

\begin{figure}[t]
          \centering
          \includegraphics[width=1\columnwidth]{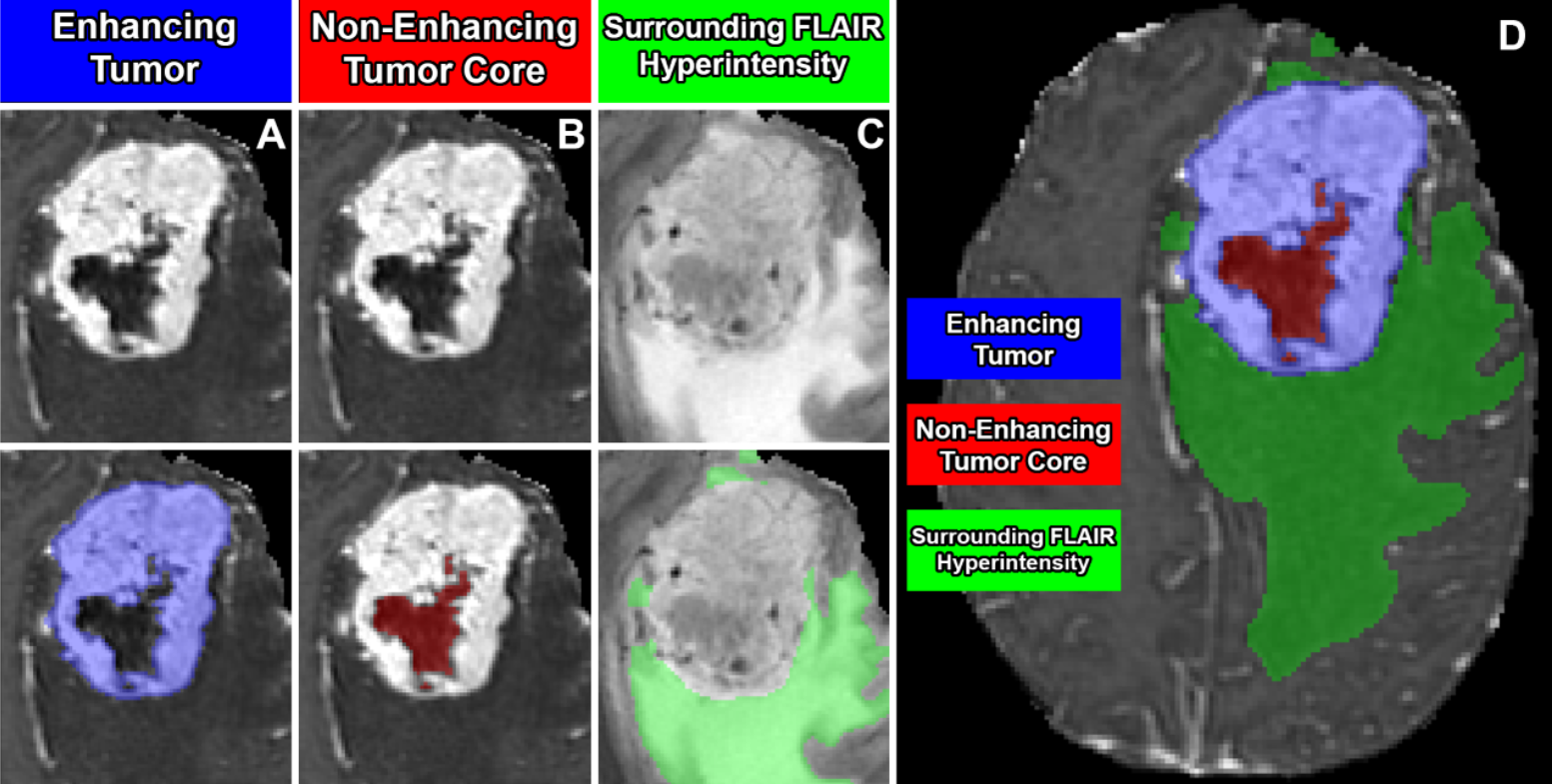}
          \caption{\textbf {Meningioma sub-regions considered in the RSNA-ASNR-MICCAI BraTS 2023 Meningioma Challenge.} Image panels with the tumor sub-regions annotated in the different mpMRI scans. The image panels A-C denote the regions considered for the performance evaluation of the participating algorithms and specifically highlight: the enhancing tumor (blue) visible in a T1Gd scan, (panel A); the non-enhancing tumor core (red) visible in a T1Gd scan, (panel B); and the surrounding FLAIR hyperintensity (green) visible in a T2-FLAIR scan. Panel D depicts the combined segmentations generating the final tumor sub-region labels, as provided to the BraTS 2023 meningioma challenge participants: enhancing tumor (blue), non-enhancing tumor core (red), and edema (green). } 
        \label{annotations}
    \end{figure}

\subsection{Data Description}    

All MRI studies in the BraTS meningioma challenge were performed in the pre-operative and pre-treatment setting and were included if one or more tumors radiographically or pathologically consistent with meningioma were included within the field of view. MRI studies containing any intracranial tumor that was not radiographically or pathologically consistent with meningioma were excluded (including cases of neurofibromatosis type 2 with intracranial Schwannomas). All cases include multiparametric MRI (mpMRI) consisting of pre-contrast T1-weighted, post-contrast T1-weighted, T2-weighted, and T2-weighted Fluid Attenuated Inversion Recovery (FLAIR) series.

\subsection{Participating sites}  

Preoperative mpMRI data for the BraTS Meningioma Challenge were contributed by academic medical centers across the United States (Table 1). Cases were identified based on histopathologic assessment following resection or biopsy or based on a formal clinical and radiographic diagnosis of meningioma, often identified based on the International Classification of Diseases Tenth Revision (ICD-10) code D32.9, "benign neoplasm of the meninges". This differs from prior BraTS glioma challenges where inclusion was based on histopathologic diagnosis alone; however, unlike high-grade gliomas, meningiomas are commonly diagnosed by imaging alone and may be observed and/or treated without a definitive tissue diagnosis. The specific case inclusion methods (pathologic, clinical/radiologic, or both) and case collection methods (i.e. retrospective, prospective, consecutive) were chosen by each participating site independently, often on the basis of pre-existing curated datasets. Imaging parameters including field strength, echo/repetition time, slice resolution, and slice thickness varied considerably between and within sites. In an effort to encourage data contribution, data contributors were not required to disclose data collection methods or MRI protocol information.
\linebreak

\begin{center}
\[
\begin{array}{ |c|c| }
\hline
\textbf{Site} & \textbf{Meningioma cases (approximate)} \\
\hline
\text{Duke University} & 450  \\ 
\text{Yale University} & 400 \\ 
\text{Thomas Jefferson} & 350 \\ 
\text{University of California San Francisco} & 200 \\ 
\text{Missouri University} & 200 \\ 
\text{University of Pennsylvania} & 50 \\ 
\text{Total} & 1650 \\
\hline
\end{array}
\]
\end{center}

Table 1: Total meningioma case contributions from each participating site. Only meningiomas that are pre-operative, within the intracranial brain mask, and radiographically or pathologically consistent with meningioma are included.

\subsection{Image data preprocessing} 

All mpMRI data underwent standardized image pre-processing steps including conversion from DICOM to Neuroimaging Informatics Technology Initiative (NIfTI) image file format; co-registration of individual image series (T1-weighted, T2-weighted, T2-FLAIR, T1Gd) to the SRI24 atlas space including uniform 1 mm$^{\text{3}}$ isotropic resampling, and automated skull-stripping using a deep convolutional neural network approach. These basic image pre-processing steps are implemented in the open-source and publicly available Federated Tumor Segmentation (FeTS) tool\footnote{\url{https://fets-ai.github.io/Front-End/}}. It should be noted that meningioma can extend through the skull and/or skull-base foramina and that any extra-cranial portions of tumors were implicitly excluded by the skull-stripping process. Despite this limitation, skull-stripping was included in the pre-processing to preserve patient anonymity (by preventing face reconstruction) and to ensure consistency with other BraTS challenges.

\subsection{Pre-segmentation} 

Prior to manual segmentation, a deep convolutional neural network-based automated segmentation model was used for automated multi-compartment pre-segmentation. This model, implemented in nnU-Net\footnote{\url{https://github.com/MIC-DKFZ/nnUNet}} was initially trained on a sample of 73 manually labeled studies from a single participating institution \cite{isensee2021nnu}. Of note, this initial sample consisted entirely of meningiomas that subsequently underwent surgical resection, which may bias the model to poorer performance for non-surgical meningiomas. During the manual correction phase of the challenge preparation, the automated segmentation algorithm was periodically retrained using additional manually corrected cases from other participating sites, including sites that contributed non-surgical meningioma cases. The purpose of iteratively retraining the model with new data was to improve its generalizability to different MRI appearances of meningioma and reduce pre-segmentation bias. Model weights for each of the different pre-segmentation models will be made publicly available at the conclusion of the challenge.

\subsection{Manual corrections} 

For each meningioma case, manual review and correction of pre-segmented labels was performed by individual volunteer “annotators” with widely varying experience levels spanning from medical students to fellowship trained neuroradiologists with 10+ years of experience. Manual corrections were performed using ITK-SNAP, a free, open-source, multi-platform software application used to segment structures in 3D and 4D biomedical images \cite{yushkevich2006user}. Annotators were provided with each of the following: 1) basic instruction on using ITK-SNAP for meningioma segmentation, 2) written descriptions of the composition of each tumor sub-compartment, and 3) a list (with examples) of common pre-segmentation errors to identify and address. These steps were designed to reduce inter-observer variability in tumor segmentation correction. After manual correction by annotators, each case was manually reviewed by a fellowship trained neuroradiologist “approver” before inclusion in the challenge dataset. In cases where an approver identified an inaccurate or incomplete segmentation, the case was returned to a different annotator for further refinement until the approver verified the manual correction.

\subsection{Common errors of automated segmentation} 

Based on subjective review of pre-segmented meningioma cases by challenge approvers, a set of commonly encountered automated segmentation errors were identified and provided to challenge annotators in an effort to improve inter-observer variability. These commonly encountered errors included:

            \begin{enumerate}
                \item  A thin rim of erroneously assigned SNFH label immediately surrounding smaller meningiomas without any true associated SNFH (Fig. 4a).
                \item  Incomplete or absent segmentation of small convexity meningiomas composed entirely of enhancing tumor, particularly when more than 1 meningioma was included in the field of view (Fig. 4b)
                \item  Improper assignment or incomplete segmentation of non-enhancing tumor regions, including exophytic hyperostosis, cystic spaces, and areas of intrinsic T1 hyperintensity, which were sometimes erroneously labeled as enhancing tumor or SNFH rather than non-enhancing tumor core (Fig. 4c)
                \item  Inclusion of non-tumor related brain parenchymal T2/FLAIR signal abnormality, most commonly chronic microvascular ischemic white matter changes (e.g. leukoaraiosis) within the SNFH label (Fig. 4d).
            \end{enumerate}

\begin{figure}[h]
          \centering
          \includegraphics[width=0.5\columnwidth]{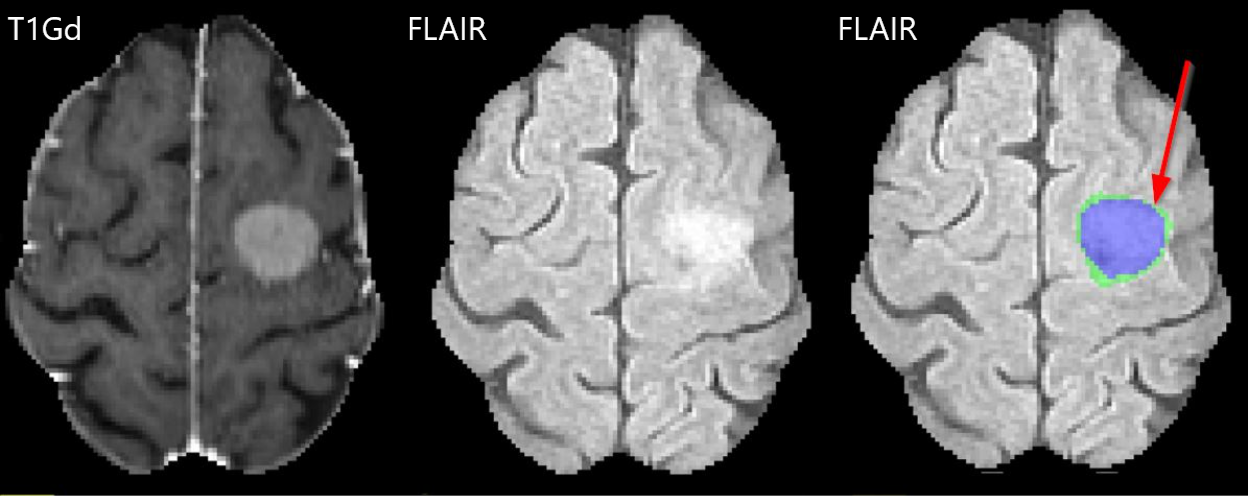} 
          \linebreak
          (a) Erroneously marked a thin rim of edema that does not exist.
          \linebreak
          \linebreak
          \includegraphics[width=0.5\columnwidth]{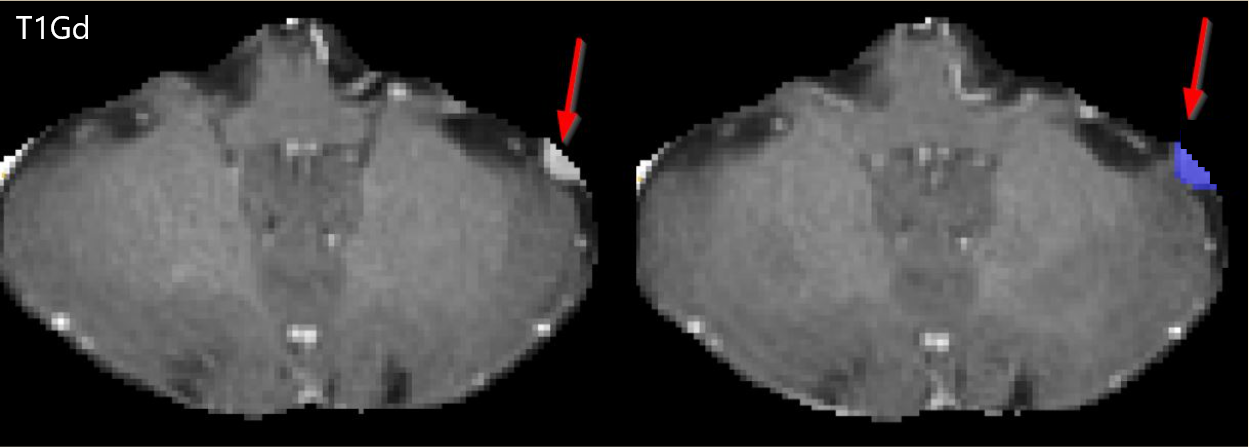}
          \linebreak
          (b) Missed small convexity meningioma.
          \linebreak
          \linebreak
          \includegraphics[width=0.5\columnwidth]{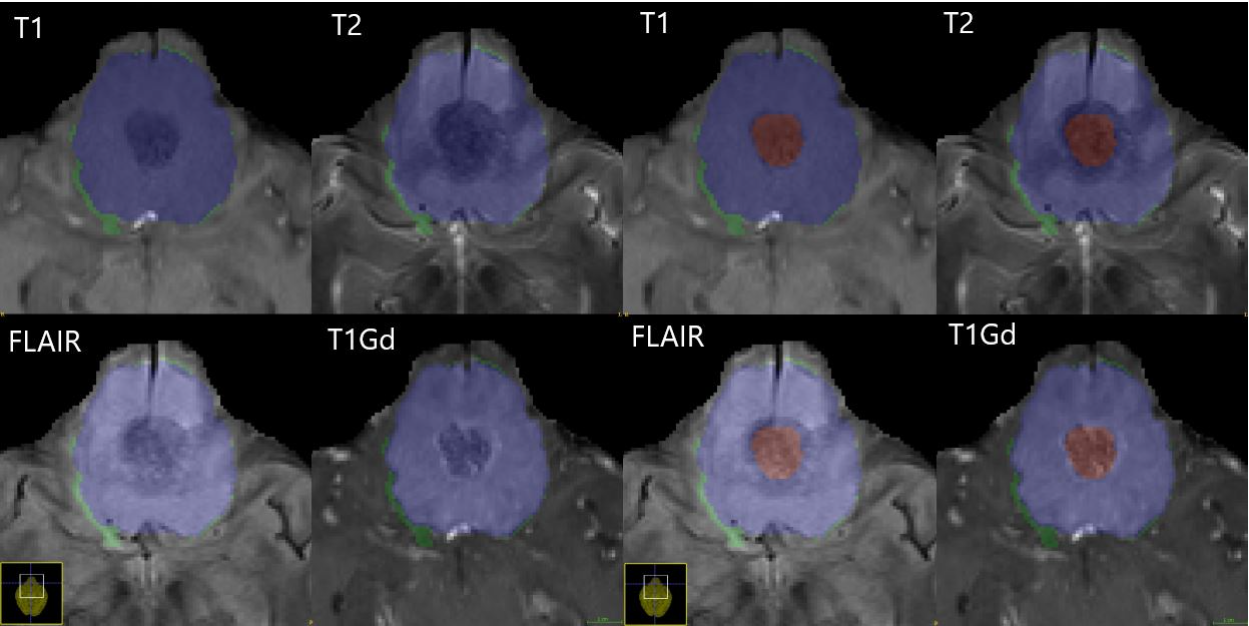}
          \linebreak
          (c) Improper classification of non-enhancing tumor.
          \linebreak
          \linebreak
          \includegraphics[width=0.5\columnwidth]{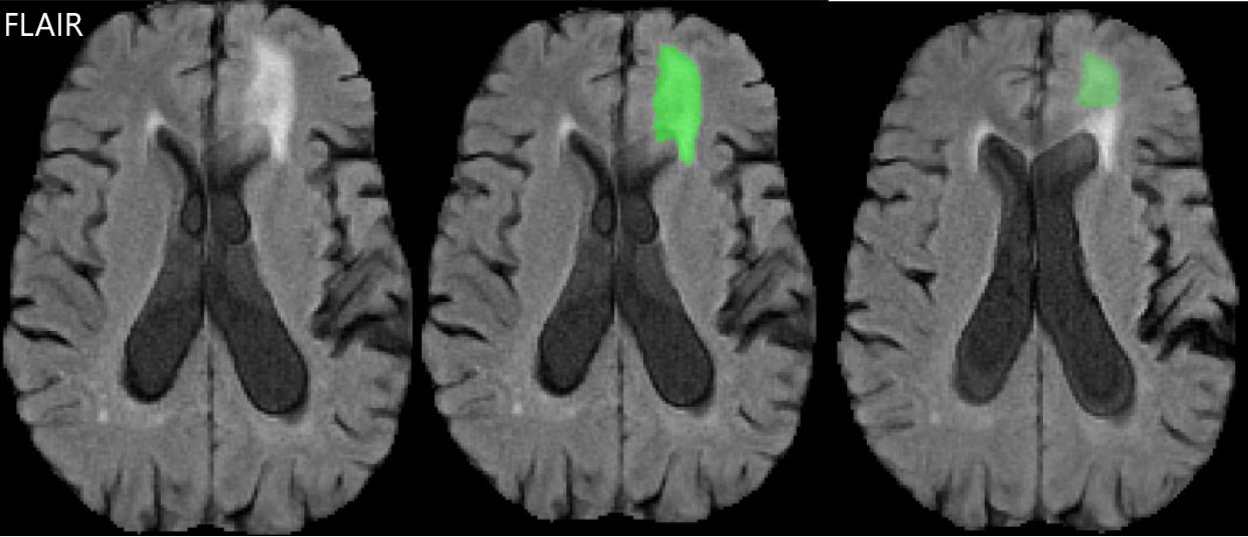}
          \linebreak
          (d) Edema bleeds into periventricular FLAIR abnormalities.
          \caption{Common errors expected from automated segmentation of meningioma sub-regions.} 
        \label{annotations}
    \end{figure}

        \section{Discussion}

\subsection{Potential benefits of the challenge}
The BraTS meningioma 2023 challenge will allow competitors to develop automated segmentation models based on the largest multi-label annotated meningioma mpMRI dataset known to date \cite{vassantachart2022automatic}. By including an expert annotated, large, multi-site, heterogenous, multi-parametric MRI meningioma dataset, there is increased variability in the challenge training dataset. This variability comes from different acquisition and processing protocols across each of the contributing sites, termed “batch effects” \cite{bento2022deep}. Having variation in the training dataset is crucial because it closely mirrors the actual underlying data distribution, which may ultimately improve the tool's overall generalizability \cite{bento2022deep}. Generalizability is a leading factor for the slow adoption of automated segmentation models into the clinical workflow. Automated segmentation models developed using large and diverse mpMRI datasets are expected to have increased performance and increased generalizability when deployed to new sites, which is in turn expected to increase likelihood of clinical adoption.

\subsection{Clinical relevance}
Accurate, reproducible, and automated meningioma segmentation models have several potential clinical uses. In the short term, automated segmentations are immediately applicable to radiation oncologists who already generate similar segmentations for radiation treatment planning. In the intermediate term, automated segmentations will be useful to radiologists for objective volumetric assessment of tumor growth on serial imaging studies and to neurosurgeons for aiding in operative planning. In the longer term, these automated segmentations will serve as the starting point for predictive models aimed at non-invasive identification of meningioma grade, subtype, aggressiveness, and response to therapy.

In radiation oncology, physicians spend significant time segmenting anatomical structures during radiation therapy planning \cite{byun2021evaluation,cha2021clinical}. Anatomical structures include gross tumor volume (GTV), clinical tumor volume (CTV) and organs-at-risk (OARs) \cite{berthelsen2007s}. Modern segmentation methods, sometimes using fused CT and MRI, have demonstrated “clinically acceptable” automated segmentation performance for several OARs and can reduce manual segmentation time by up to 43\% \cite{vrtovec2020auto,movcnik2018segmentation,ginn2023clinical}. “Clinical acceptability” as defined by Baroudi et al roughly corresponds to a Dice overlap coefficient of $>$ 0.90 for automated segmentation compared to manual segmentation, which often corresponds with inter-observer agreement of determining a clinically acceptable segmentation volume \cite{baroudi2023automated}. However, despite success with automated segmentation of OARs, GTV segmentation models have yet to be widely implemented into clinical practice. State-of-the-art deep learning models trained on diverse multi-institution data, like those that will be generated by the BraTS meningioma 2023 challenge, have the potential to approach or exceed the “clinically acceptable” range for GTV segmentation and therefore may be immediately applicable to clinical radiation planning.

In image guided neurosurgery, surgeons utilize co-registered preoperative anatomical, functional, and diffusion tensor imaging to assist with tumor localization intraoperatively \cite{drakopoulos2021adaptive}. Automated segmentations could provide neurosurgeons with additional information to aid in surgical planning, particularly with respect to determining tumor proximity to eloquent white matter tracts identified by DTI tractography and eloquent cerebral cortex identified by functional MRI.

In radiology, automated segmentation has had a growing role in recent years. For example, Lee et al. describe a deep learning algorithm that computes longitudinal volume measurement of vestibular schwannomas following Gamma Knife radiosurgery (GKRS) \cite{lee2021applying}. The model described by Lee had an average of -0.44\% to +1.76\% relative volume difference between the AI prediction and expert radiologist measurement over 861 patients in follow-up, which highlights the potential role for automated segmentation methods in longitudinal volumetric assessment. Rudie et al. trained a multichannel multiclass segmentation nnU-Net model on posttreatment diffuse gliomas as well as a separate longitudinal change nnU-Net model to help localize and quantify changes in treatment change tissue and active tumor \cite{rudie2022longitudinal}. They found that the accuracy levels of the longitudinal change networks to predict an increase, decrease, or no change in peritumoral edema or active tumor volumes were not significantly different from those of three neuroradiologists. By automatically computing the volume of tumor at follow-up appointments, providers can more accurately determine progression, pseudo-progression, or regression of the treated disease.  In addition, automated segmentations have been used for radiogenomic studies aimed at non-invasive prediction of clinically relevant tumor characteristics such as genetic subtypes of gliomas \cite{buda2020deep,calabrese2022combining}. One retrospective study analyzed pre-operative MRIs from patients with WHO grade IV diffuse astrocytic gliomas and showed that combining radiomics with CNN features improved genetic biomarker prediction \cite{calabrese2022combining}. To date, there has been relatively little investigation into radiogenomics for meningiomas. Clinically, the WHO grade of a meningioma is one of the main determinants of treatment recommendations, and determination by resection or biopsy is not always desirable. If a deep radiogenomics model could accurately identify subtypes, and therefore prognosticate a meningioma prior to intervention, then patients could be spared the risks associated with biopsy, resection, or radiation therapy.

\subsection{Limitations of the challenge}
While great care and attention was provided during challenge design and dataset preparation, there are several important limitations that should be addressed. First, pre-processing of the mpMRI images introduces loss of radiographic information that would otherwise be analyzed in clinical practice. For example, skull-stripping implicitly removes any portion of a meningioma that extends beyond the confines of the cranial vault, which limits potential clinical utility for meningiomas that are not entirely intracranial. In addition, series co-registration and resampling to a standardized atlas space obscures the native image acquisition resolution and inherently changes the image data through interpolation. These pre-processing steps also introduce an additional step (and therefore potential barrier) to clinical implementation of automated tumor segmentation algorithms. These limitations, while significant, were deemed necessary in order to protect patient privacy, to remain consistent with other BraTS challenges, and to lower potential data-related barriers to challenge participation. Second, the annotator-approval model for manual segmentation corrections, while consistent with prior BraTS challenges, does not address the issue of inter-observer variability, which could fundamentally limit the performance of any model trained on these data. Of specific concern with regards to meningioma, dural tail enhancement segmentation has significant inter-observer variability due to close proximity to normal enhancing structures (e.g. normal dura, dural vasculature) as well as historically ill-defined definitions of what constitutes a dural tail. In order to reduce this variability, annotators were provided with specific written instructions and examples of commonly encountered pre-segmentation errors, and an expert approver reviewed all segmentations before inclusion in the challenge dataset. Ultimately, expert clinical judgment on meningioma boundaries will likely never be perfectly aligned between different observers, and these issues will only be fully addressed by the labor- and time-intensive process of multi-observer consensus segmentation.

\subsection{Goals for future challenges}

There are many potential ways to expand and improve the BraTS meningioma challenge in future iterations. Perhaps the most straightforward extension would be using the current challenge dataset to predict tumor WHO grade, which would not require any new imaging data and could provide substantial clinical utility for determination of prognosis and optimal treatment course. Existing data could also be used to detect (rather than merely segment), localize, and quantify (with linear or volumetric measurements) intracranial meningiomas, which would be useful for improving radiologist’s workflows and improving objectivity in the longitudinal follow up setting.

Several additional potential extensions of the challenge would require new inclusion/exclusion criteria and additional data, but also offer improved clinical utility. For example, while the BraTS meningioma 2023 challenge is focused on preoperative imaging, meningioma imaging assessment and treatment planning are frequently performed in the postoperative and/or post-treatment setting. In addition, while most meningiomas are predominantly intracranial, extracranial meningiomas involving the face, skull-base, or spinal canal are frequently encountered in clinical practice. Future BraTS meningioma and related challenges should incorporate non-skullstripped mpMRI images of previously treated meningiomas to allow development of automated segmentation models that could be clinically useful for a wider variety of patients with meningioma. 

Another important clinical dilemma that could be addressed by future challenges is determination of recurrence versus treatment related changes in patients who have undergone definitive surgical resection and/or radiotherapy for meningioma. Non-pathologic treatment related changes, especially following radiation therapy, have overlapping radiographic findings with recurrent meningioma. Therefore, developing a model to help distinguish recurrence from treatment changes could provide substantial clinical utility for developing optimal patient management plans.

        \section{Conclusion}
The BraTS meningioma 2023 challenge will provide a community standard and benchmark for automated intracranial meningioma segmentation models based on the largest expert annotated multilabel mpMRI image dataset to date. The state-of-the-art models developed during the course of this competition will lead to clinically viable automated meningioma segmentation methods that can aid in objective tumor monitoring, surgical and radiotherapy planning, and non-invasive assessments of clinically relevant tumor characteristics that will ultimately improve care of patients with meningiomas.

    \section*{Acknowledgments}
        Developing large and well curated mpMRI datasets for auto-segmentation model development requires significant time and expertise from neuro-radiology experts. We are grateful to everyone who contributed to the development and review of the tumor volume labels including volunteer annotators/approvers from the American Society of Neuroradiology (ASNR).
    
    \section*{Funding}
    
    Research reported in this publication was partly supported by the National Institutes of Health (NIH) under award numbers: NCI K08CA256045 and NCI/ITCR U01CA242871. The content of this publication is solely the responsibility of the authors and does not represent the official views of the NIH.
        
    \bibliographystyle{ieeetr}
    \bibliography{bibliography.bib}

\begin{thebibliography}{10}

\bibitem{ogasawara2021meningioma}
C.~Ogasawara, B.~D. Philbrick, and D.~C. Adamson, ``Meningioma: a review of
  epidemiology, pathology, diagnosis, treatment, and future directions,'' {\em
  Biomedicines}, vol.~9, no.~3, p.~319, 2021.

\bibitem{huntoon2020meningioma}
K.~Huntoon, A.~M.~S. Toland, and S.~Dahiya, ``Meningioma: a review of
  clinicopathological and molecular aspects,'' {\em Frontiers in Oncology},
  vol.~10, p.~579599, 2020.

\bibitem{khanna2021machine}
O.~Khanna, A.~F. Kazerooni, C.~J. Farrell, M.~P. Baldassari, T.~D. Alexander,
  M.~Karsy, B.~A. Greenberger, J.~A. Garcia, C.~Sako, J.~J. Evans, {\em
  et~al.}, ``Machine learning using multiparametric magnetic resonance imaging
  radiomic feature analysis to predict ki-67 in world health organization grade
  i meningiomas,'' {\em Neurosurgery}, vol.~89, no.~5, pp.~928--936, 2021.

\bibitem{bakas2018identifying}
S.~Bakas, M.~Reyes, A.~Jakab, S.~Bauer, M.~Rempfler, A.~Crimi, R.~Shinohara,
  C.~Berger, S.~Ha, M.~Rozycki, {\em et~al.}, ``Identifying the best machine
  learning algorithms for brain tumor segmentation,'' {\em Progression
  Assessment, and Overall Survival Prediction in the BRATS Challenge}, vol.~10,
  2018.

\bibitem{menze2014multimodal}
B.~H. Menze, A.~Jakab, S.~Bauer, J.~Kalpathy-Cramer, K.~Farahani, J.~Kirby,
  Y.~Burren, N.~Porz, J.~Slotboom, R.~Wiest, {\em et~al.}, ``The multimodal
  brain tumor image segmentation benchmark (brats),'' {\em IEEE transactions on
  medical imaging}, vol.~34, no.~10, pp.~1993--2024, 2014.

\bibitem{bakas2017advancing}
S.~Bakas, H.~Akbari, A.~Sotiras, M.~Bilello, M.~Rozycki, J.~S. Kirby, J.~B.
  Freymann, K.~Farahani, and C.~Davatzikos, ``Advancing the cancer genome atlas
  glioma mri collections with expert segmentation labels and radiomic
  features,'' {\em Scientific data}, vol.~4, no.~1, pp.~1--13, 2017.

\bibitem{bakas2017segmentation_1}
S.~Bakas, H.~Akbari, A.~Sotiras, {\em et~al.}, ``Segmentation labels for the
  pre-operative scans of the tcga-gbm collection.,'' {\em The cancer imaging
  archive}, 2017.

\bibitem{bakas2017segmentation_2}
S.~Bakas, H.~Akbari, A.~Sotiras, M.~Bilello, M.~Rozycki, J.~Kirby, J.~Freymann,
  K.~Farahani, and C.~Davatzikos, ``Segmentation labels and radiomic features
  for the pre-operative scans of the tcga-lgg collection,'' {\em The cancer
  imaging archive}, vol.~286, 2017.

\bibitem{bakas2022university}
S.~Bakas, C.~Sako, H.~Akbari, M.~Bilello, A.~Sotiras, G.~Shukla, J.~D. Rudie,
  N.~F. Santamar{\'\i}a, A.~F. Kazerooni, S.~Pati, {\em et~al.}, ``The
  university of pennsylvania glioblastoma (upenn-gbm) cohort: Advanced mri,
  clinical, genomics, \& radiomics,'' {\em Scientific data}, vol.~9, no.~1,
  p.~453, 2022.

\bibitem{baid2021rsna}
U.~Baid, S.~Ghodasara, S.~Mohan, M.~Bilello, E.~Calabrese, E.~Colak,
  K.~Farahani, J.~Kalpathy-Cramer, F.~C. Kitamura, S.~Pati, {\em et~al.}, ``The
  rsna-asnr-miccai brats 2021 benchmark on brain tumor segmentation and
  radiogenomic classification,'' {\em arXiv preprint arXiv:2107.02314}, 2021.

\bibitem{baid2020novel}
U.~Baid, S.~Talbar, S.~Rane, S.~Gupta, M.~H. Thakur, A.~Moiyadi, N.~Sable,
  M.~Akolkar, and A.~Mahajan, ``A novel approach for fully automatic
  intra-tumor segmentation with 3d u-net architecture for gliomas,'' {\em
  Frontiers in computational neuroscience}, p.~10, 2020.

\bibitem{martz2022anocef}
N.~Martz, J.~Salleron, F.~Dhermain, G.~Vogin, J.~Daisne, R.~M. Audouard,
  R.~Tanguy, G.~Noel, M.~Peyre, I.~Lecouillard, {\em et~al.}, ``Anocef
  consensus guideline on target volume delineation for meningiomas
  radiotherapy,'' {\em International Journal of Radiation Oncology, Biology,
  Physics}, vol.~114, no.~3, p.~e46, 2022.

\bibitem{nasa2021delphi}
P.~Nasa, R.~Jain, and D.~Juneja, ``Delphi methodology in healthcare research:
  how to decide its appropriateness,'' {\em World Journal of Methodology},
  vol.~11, no.~4, p.~116, 2021.

\bibitem{watts2014magnetic}
J.~Watts, G.~Box, A.~Galvin, P.~Brotchie, N.~Trost, and T.~Sutherland,
  ``Magnetic resonance imaging of meningiomas: a pictorial review,'' {\em
  Insights into imaging}, vol.~5, no.~1, pp.~113--122, 2014.

\bibitem{bitzer1998angiogenesis}
M.~Bitzer, H.~Opitz, J.~Popp, M.~Morgalla, A.~Gruber, E.~Heiss, and K.~Voigt,
  ``Angiogenesis and brain oedema in intracranial meningiomas: influence of
  vascular endothelial growth factor,'' {\em Acta neurochirurgica}, vol.~140,
  pp.~333--340, 1998.

\bibitem{o2007atypical}
S.~O'leary, W.~Adams, R.~Parrish, and W.~Mukonoweshuro, ``Atypical imaging
  appearances of intracranial meningiomas,'' {\em Clinical radiology}, vol.~62,
  no.~1, pp.~10--17, 2007.

\bibitem{dorent2023crossmoda}
R.~Dorent, A.~Kujawa, M.~Ivory, S.~Bakas, N.~Rieke, S.~Joutard, B.~Glocker,
  J.~Cardoso, M.~Modat, K.~Batmanghelich, {\em et~al.}, ``Crossmoda 2021
  challenge: Benchmark of cross-modality domain adaptation techniques for
  vestibular schwannoma and cochlea segmentation,'' {\em Medical Image
  Analysis}, vol.~83, p.~102628, 2023.

\bibitem{isensee2021nnu}
F.~Isensee, P.~F. Jaeger, S.~A. Kohl, J.~Petersen, and K.~H. Maier-Hein,
  ``nnu-net: a self-configuring method for deep learning-based biomedical image
  segmentation,'' {\em Nature methods}, vol.~18, no.~2, pp.~203--211, 2021.

\bibitem{yushkevich2006user}
P.~A. Yushkevich, J.~Piven, H.~C. Hazlett, R.~G. Smith, S.~Ho, J.~C. Gee, and
  G.~Gerig, ``User-guided 3d active contour segmentation of anatomical
  structures: significantly improved efficiency and reliability,'' {\em
  Neuroimage}, vol.~31, no.~3, pp.~1116--1128, 2006.

\bibitem{vassantachart2022automatic}
A.~Vassantachart, Y.~Cao, M.~Gribble, S.~Guzman, J.~C. Ye, K.~Hurth, A.~Mathew,
  G.~Zada, Z.~Fan, E.~L. Chang, {\em et~al.}, ``Automatic differentiation of
  grade i and ii meningiomas on magnetic resonance image using an asymmetric
  convolutional neural network,'' {\em Scientific Reports}, vol.~12, no.~1,
  p.~3806, 2022.

\bibitem{bento2022deep}
M.~Bento, I.~Fantini, J.~Park, L.~Rittner, and R.~Frayne, ``Deep learning in
  large and multi-site structural brain mr imaging datasets,'' {\em Frontiers
  in Neuroinformatics}, vol.~15, p.~82, 2022.

\bibitem{byun2021evaluation}
H.~K. Byun, J.~S. Chang, M.~S. Choi, J.~Chun, J.~Jung, C.~Jeong, J.~S. Kim,
  Y.~Chang, S.~Y. Chung, S.~Lee, {\em et~al.}, ``Evaluation of deep
  learning-based autosegmentation in breast cancer radiotherapy,'' {\em
  Radiation Oncology}, vol.~16, no.~1, pp.~1--8, 2021.

\bibitem{cha2021clinical}
E.~Cha, S.~Elguindi, I.~Onochie, D.~Gorovets, J.~O. Deasy, M.~Zelefsky, and
  E.~F. Gillespie, ``Clinical implementation of deep learning contour
  autosegmentation for prostate radiotherapy,'' {\em Radiotherapy and
  Oncology}, vol.~159, pp.~1--7, 2021.

\bibitem{berthelsen2007s}
A.~K. Berthelsen, J.~Dobbs, E.~Kjell{\'e}n, T.~Landberg, T.~R. M{\"o}ller,
  P.~Nilsson, L.~Specht, and A.~Wambersie, ``What's new in target volume
  definition for radiologists in icru report 71? how can the icru volume
  definitions be integrated in clinical practice?,'' {\em Cancer Imaging},
  vol.~7, no.~1, p.~104, 2007.

\bibitem{vrtovec2020auto}
T.~Vrtovec, D.~Mo{\v{c}}nik, P.~Strojan, F.~Pernu{\v{s}}, and B.~Ibragimov,
  ``Auto-segmentation of organs at risk for head and neck radiotherapy
  planning: from atlas-based to deep learning methods,'' {\em Medical physics},
  vol.~47, no.~9, pp.~e929--e950, 2020.

\bibitem{movcnik2018segmentation}
D.~Mo{\v{c}}nik, B.~Ibragimov, L.~Xing, P.~Strojan, B.~Likar, F.~Pernu{\v{s}},
  and T.~Vrtovec, ``Segmentation of parotid glands from registered ct and mr
  images,'' {\em Physica Medica}, vol.~52, pp.~33--41, 2018.

\bibitem{ginn2023clinical}
J.~S. Ginn, H.~A. Gay, J.~Hilliard, J.~Shah, N.~Mistry, C.~M{\"o}hler, G.~D.
  Hugo, and Y.~Hao, ``A clinical and time savings evaluation of a deep learning
  automatic contouring algorithm,'' {\em Medical Dosimetry}, vol.~48, no.~1,
  pp.~55--60, 2023.

\bibitem{baroudi2023automated}
H.~Baroudi, K.~K. Brock, W.~Cao, X.~Chen, C.~Chung, L.~E. Court, M.~D.
  El~Basha, M.~Farhat, S.~Gay, M.~P. Gronberg, {\em et~al.}, ``Automated
  contouring and planning in radiation therapy: What is ‘clinically
  acceptable’?,'' {\em Diagnostics}, vol.~13, no.~4, p.~667, 2023.

\bibitem{drakopoulos2021adaptive}
F.~Drakopoulos, C.~Tsolakis, A.~Angelopoulos, Y.~Liu, C.~Yao, K.~R. Kavazidi,
  N.~Foroglou, A.~Fedorov, S.~Frisken, R.~Kikinis, {\em et~al.}, ``Adaptive
  physics-based non-rigid registration for immersive image-guided
  neuronavigation systems,'' {\em Frontiers in Digital Health}, vol.~2,
  p.~613608, 2021.

\bibitem{lee2021applying}
C.-c. Lee, W.-K. Lee, C.-C. Wu, C.-F. Lu, H.-C. Yang, Y.-W. Chen, W.-Y. Chung,
  Y.-S. Hu, H.-M. Wu, Y.-T. Wu, {\em et~al.}, ``Applying artificial
  intelligence to longitudinal imaging analysis of vestibular schwannoma
  following radiosurgery,'' {\em Scientific reports}, vol.~11, no.~1, p.~3106,
  2021.

\bibitem{rudie2022longitudinal}
J.~D. Rudie, E.~Calabrese, R.~Saluja, D.~Weiss, J.~B. Colby, S.~Cha, C.~P.
  Hess, A.~M. Rauschecker, L.~P. Sugrue, and J.~E. Villanueva-Meyer,
  ``Longitudinal assessment of posttreatment diffuse glioma tissue volumes with
  three-dimensional convolutional neural networks,'' {\em Radiology: Artificial
  Intelligence}, vol.~4, no.~5, p.~e210243, 2022.

\bibitem{buda2020deep}
M.~Buda, E.~A. AlBadawy, A.~Saha, and M.~A. Mazurowski, ``Deep radiogenomics of
  lower-grade gliomas: Convolutional neural networks predict tumor genomic
  subtypes using mr images,'' {\em Radiology: Artificial Intelligence}, vol.~2,
  no.~1, p.~e180050, 2020.

\bibitem{calabrese2022combining}
E.~Calabrese, J.~D. Rudie, A.~M. Rauschecker, J.~E. Villanueva-Meyer, J.~L.
  Clarke, D.~A. Solomon, and S.~Cha, ``Combining radiomics and deep
  convolutional neural network features from preoperative mri for predicting
  clinically relevant genetic biomarkers in glioblastoma,'' {\em Neuro-Oncology
  Advances}, vol.~4, no.~1, p.~vdac060, 2022.

\end{thebibliography}
    \newpage
    \appendix
\end{document}